%% file: cvpr_2017_depth_estimation_final.tex
\documentclass[10pt,twocolumn,letterpaper]{article}

\usepackage{cvpr}
\usepackage{times}
\usepackage{epsfig}
\usepackage{graphicx}
\usepackage{amsmath}
\usepackage{booktabs}
\usepackage{amssymb}
\usepackage {multirow}
\usepackage{fixltx2e}

\newcommand{\vect}[1]{\mathbf{#1}}

\newcommand{\mr}[1]{\mathrm{#1}}

\def\eg{\textit{e.g.}~}
\def\ie{\textit{i.e.}~}

\def\etal{\textit{et al.}~}

\usepackage{tabularx}

\newcommand{\tabincell}[2]{\begin{tabular}{@{}#1@{}}#2\end{tabular}}

\usepackage[breaklinks=true,bookmarks=false]{hyperref}

\cvprfinalcopy 

\def\cvprPaperID{2250} 

\ifcvprfinal\fi
\begin{document}
\title{Multi-Scale Continuous CRFs as Sequential Deep Networks \\for Monocular Depth Estimation}

\author{Dan Xu$^1$, Elisa Ricci$^{4,5}$, Wanli Ouyang$^{2,3}$, Xiaogang Wang$^2$, Nicu Sebe$^1$\\
$^1$University of Trento, $^{2}$The Chinese University of Hong Kong\\
$^3$The University of Sydney, $^4$Fondazione Bruno Kessler, $^5$University of Perugia\\
{\tt\small \{dan.xu, niculae.sebe\}@unitn.it, eliricci@fbk.eu, \{wlouyang, xgwang\}@ee.cuhk.edu.hk}  
}

\maketitle
  \newcommand\figcaption{\def\@captype{figure}\caption} 
  \newcommand\tabcaption{\def\@captype{table}\caption} 


\begin{abstract}
This paper addresses the problem of depth estimation from a single still image.
Inspired by recent works on multi-scale convolutional neural networks (CNN), we propose a 
deep model which fuses complementary 
information derived from multiple CNN side outputs.
Different from previous methods, 
the integration is obtained by means of 
continuous Conditional Random Fields (CRFs). In particular, we propose {two different variations}, 
one based on a cascade of multiple CRFs, the other on a unified graphical model. By designing a novel CNN
implementation of mean-field updates for continuous CRFs, we show that both proposed models
can be regarded as sequential deep networks and that 
training can be performed end-to-end. 
Through extensive experimental evaluation we demonstrate the effectiveness of the proposed approach 
and establish new state of the art results on publicly available datasets. 
\end{abstract}

\section{Introduction}
\input{introduction.tex}

\section{Related work}
\input{relatedwork.tex}

\section{Multi-Scale Models for Depth Estimation}
\input{approach.tex}

\section{Experiments}
\input{experiments_e}
\vspace{-0.1cm}

\section{Conclusions}
We introduced a novel approach for predicting depth images from a single RGB input, which is also particularly useful for other cross-modal tasks~\cite{xu2013novel,xu2017cvpr}. The core of the method is a novel framework based on continuous CRFs for fusing multi-scale representations derived from CNN side outputs. 
We demonstrated that this framework can be used in combination with several common CNN architectures and is suitable for end-to-end training. The extensive experiments confirmed the validity of the proposed multi-scale fusion approach. While this paper specifically addresses the problem of depth prediction, we believe that other tasks in computer vision involving pixel-level predictions
of continuous variables, can also benefit from our implementation of mean-fields updates within the CNN framework.

{\small
\bibliographystyle{ieee}
\bibliography{egbib}
}

\end{document}

%% file: introduction.tex
\begin{figure}[t]
\centering
\includegraphics[width=3.18in]{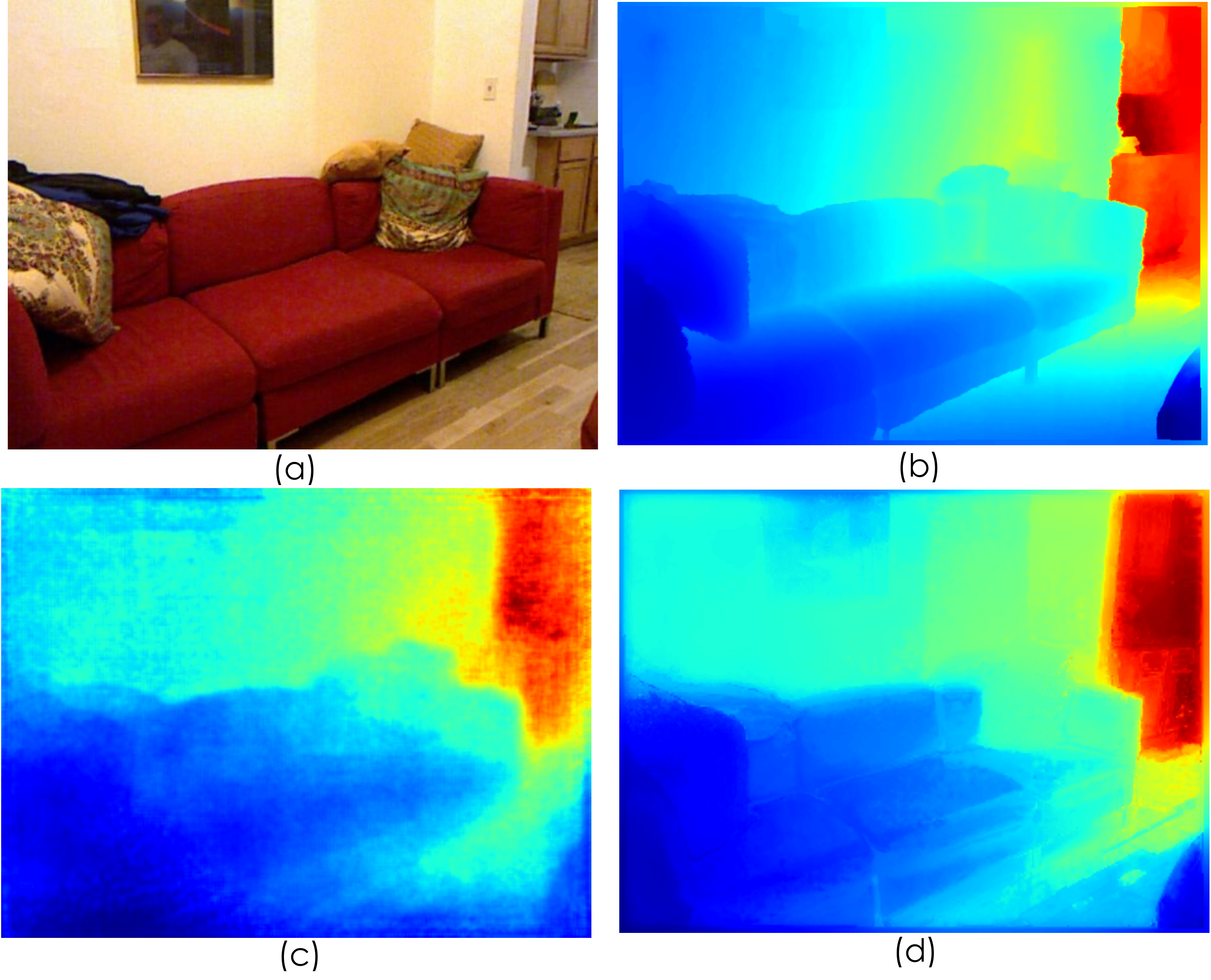}
\caption{(a) Original RGB image. (b) Ground truth. Depth map obtained by considering a pre-trained CNN (\eg VGG 
Convolution-Deconvolution \cite{noh2015learning})
and fusing multi-layer representations (c) with the approach in
\cite{xie2015holistically} and (d) with the proposed multi-scale CRF. 
}
\label{motivation}
\vspace{-0.5cm}
\end{figure}

While estimating the depth of a scene from
a single image is a natural ability for humans, devising computational models for 
accurately predicting depth information from RGB data is a challenging task.
Many attempts have been made to address this problem in the past. In particular, recent works have achieved 
remarkable performance thanks to powerful deep learning models \cite{eigen2015predicting,eigen2014depth,liu2015deep,porzi2017learning}. Assuming the 
availability of a large training set of RGB-depth pairs, 
monocular depth prediction is casted as a pixel-level regression problem and 
Convolutional Neural Network (CNN) architectures are typically employed.

In the last few years significant effort have been made in the research community 
to improve the performance of CNN models for pixel-level prediction tasks (\eg semantic segmentation, contour detection).
Previous works have shown that, for depth estimation as well as for other pixel-level classification/regression problems, more accurate estimates can be obtained by combining information from multiple scales \cite{eigen2015predicting,xie2015holistically,chen2015attention}. 
This can be achieved in different ways, \eg fusing feature maps corresponding to different network layers or 
designing an architecture with multiple inputs corresponding to images at different resolutions. 
Other works have demonstrated that, by adding a Conditional Random Field (CRF) in cascade to a convolutional neural architecture, the performance 
can be greatly enhanced and the CRF can be fully integrated within the deep model enabling end-to-end training 
with back-propagation \cite{zheng2015conditional}. However, these works mainly 
focus on pixel-level prediction problems in the discrete domain (\eg semantic segmentation).
While complementary, so far these strategies have been only considered in isolation and 
no previous works have exploited multi-scale information within a CRF inference framework. 

In this paper we argue that, benefiting from the flexibility and the representational power of graphical models, we
can optimally fuse representations derived from multiple CNN side output layers, improving
performance over traditional multi-scale strategies.
By exploiting this idea, we introduce a novel framework
to estimate depth maps from single still images. 
{Opposite to previous work fusing multi-scale features by averaging or concatenation, we propose to integrate 
multi-layer side output information by devising a novel approach based on continuous CRFs.}
Specifically, we present two different methods. The first approach is based on 
a single multi-scale CRF model, while the other considers a cascade of scale-specific CRFs. We also show that, 
by introducing a common CNN implementation for mean-fields updates in continuous CRFs, both models are equivalent to 
sequential deep networks and an end-to-end approach can be devised for training. 
Through extensive experimental evaluation
we demonstrate that the proposed CRF-based method produces more accurate depth maps than
traditional multi-scale approaches for pixel-level prediction tasks \cite{hariharan2015hypercolumns,xie2015holistically} (Fig.\ref{motivation}). Moreover,
by performing experiments on the publicly available NYU Depth V2 \cite{silberman2012indoor} and on the 
Make3D \cite{saxena2009make3d} datasets, we show that our approach outperforms state of the art methods for monocular depth estimation.

To summarize, the contributions of this paper are threefold. First, we propose a novel approach for predicting depth maps from RGB inputs
which exploits multi-scale estimations derived from CNN inner layers by fusing them within a CRF framework. 
Second, as the task of
pixel-level depth prediction implies inferring a set of continuous values, we show how
mean field (MF) updates can be implemented as sequential deep models, enabling end-to-end training of the whole network. We
believe that our MF implementation will be useful not only to researchers working on depth prediction, but also to
those interested in other problems involving continuous variables. Therefore, our code is made publicly available\footnote{https://github.com/danxuhk/ContinuousCRF-CNN.git}.
Third, our experiments demonstrate that
the proposed multi-scale CRF framework is superior to previous methods integrating information from intermediate network layers by combining multiple
losses \cite{xie2015holistically} or by adopting feature 
concatenations \cite{hariharan2015hypercolumns}. We also show that our approach outperforms state of the art depth estimation methods on 
public benchmarks and that
the proposed CRF-based models can be employed in combination with different pre-trained CNN architectures, consistently enhancing their performance.

%% file: relatedwork.tex
\paragraph{Depth Estimation.} Previous approaches for depth estimation from single images can be categorized into three main 
groups: (i) methods operating on hand crafted features, (ii) methods based on 
graphical models and (iii) methods adopting deep networks.

Earlier works addressing the depth prediction task belong to the first category.
Hoiem \etal \cite{hoiem2005automatic} introduced photo pop-up, a 
fully automatic method for creating a basic 3D model from a single photograph. 
Karsch \etal \cite{karsch2014depth} developed Depth Transfer, a non parametric approach 
where
the depth of an input image is reconstructed by transferring the depth of multiple similar images and then applying some
warping and optimizing procedures. Ladicky \cite{ladicky2014pulling} demonstrated the benefit of combining semantic object 
labels with depth features.
\vspace{-0.05cm}
\par Other works exploited the flexibility of graphical models to reconstruct depth information.
For instance, Delage \etal \cite{delage2006dynamic} proposed a dynamic Bayesian
framework for recovering 3D information from indoor scenes. A discriminatively-trained 
multiscale Markov Random Field (MRF) was introduced in \cite{saxena20083}, in order to optimally fuse 
local and global features. Depth estimation was treated as an inference problem in a 
discrete-continuous CRF in \cite{liu2014discrete}. However, these works did not 
employ deep networks.
\vspace{-0.05cm}
\par More recent approaches for depth estimation are based on CNNs \cite{eigen2015predicting,liu2015deep,wang2015towards,roymonocular,laina2016deeper}.
For instance, Eigen \etal \cite{eigen2014depth} proposed a multi-scale approach for depth prediction,
considering two deep networks, one performing a coarse global prediction
based on the entire image, and the other refining predictions locally. This approach
was extended in \cite{eigen2015predicting} to handle multiple tasks (\eg semantic segmentation,
surface normal estimation). Wang \etal \cite{wang2015towards} introduced a CNN for joint depth estimation 
and semantic segmentation. The obtained estimates were further refined with a Hierarchical CRF. 
The most similar work to ours is \cite{liu2015deep}, where the 
representational power of deep CNN and continuous CRF is jointly exploited for depth prediction.
However, the method proposed in \cite{liu2015deep} is based on superpixels and the
information associated to multiple scales is not exploited. \vspace{-3mm} 
\vspace{-0.3cm}
\paragraph{Multi-scale CNNs.} The problem of combining informations from multiple scales for pixel-level prediction
tasks have received considerable interest lately. In \cite{xie2015holistically} a deeply supervised fully convolutional neural 
network was proposed for edge detection. Skip-layer networks, where the feature maps derived from different
levels of a primary network are jointly considered in an output layer, have also become very popular
\cite{long2015fully,bertasius2015deepedge}. Other works considered multi-stream architectures, where
multiple parallel networks receiving inputs at different scale are fused \cite{buyssens2012multiscale}. Dilated
convolutions (\eg \textit{dilation} or \textit{\`{a} trous}) have been also employed in different deep network models in order to 
aggregate multi-scale contextual information \cite{chen2014semantic}.
We are not aware of previous works exploiting multi-scale representations into a continuous CRF framework.

%% file: approach.tex
\begin{figure*}[t]
\centering
\includegraphics[width=6in, height=2.6in]{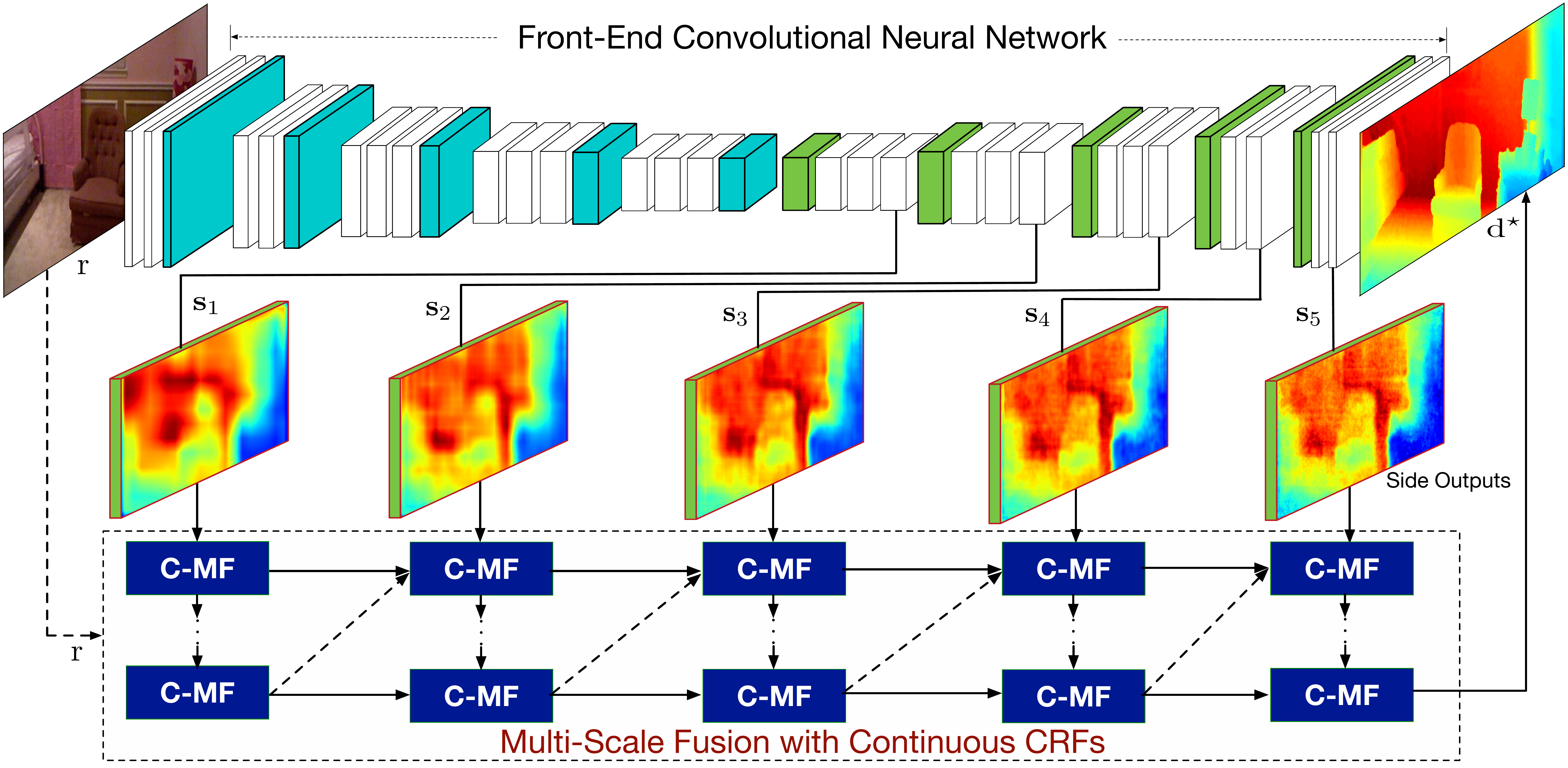} 
\caption{Overview of the proposed deep architecture. Our model is composed of two main components: a front-end CNN and 
a fusion module. The fusion module uses continuous CRFs to integrate multiple side output maps of the front-end CNN. 
We consider two different CRFs-based multi-scale models and implement them as sequential deep networks by 
stacking several elementary blocks, the C-MF blocks. 
}
\vspace{-0.4cm}
\label{framework}
\end{figure*}

In this section we introduce our approach 
for depth estimation from single images. We first formalize the problem of depth prediction. Then, we
describe two variations of the proposed multi-scale model, one based on a cascade of CRFs and the other on a 
single multi-scale CRFs. Finally, we show how our whole deep network can be trained end-to-end, introducing a 
novel CNN implementation for mean-field iterations in continuous CRFs.


\subsection{Problem Formulation and Overview}
Following previous works we formulate the task of depth prediction from monocular RGB input as the problem of learning a non-linear mapping 
$F:\mathcal{I} \rightarrow \mathcal{D}$ from the image space $\mathcal{I}$ to the output depth space $\mathcal{D}$. 
More formally, let $\mathcal{Q} = \{ (\vect{r}_i, \bar{\vect{d}}_i)\}_{i=1}^Q$ be a training set of $Q$ pairs, 
where $\vect{r}_i \in~\mathcal{I}$ denotes an input RGB image with $N$ pixels and $\bar{\vect{d}}_i \in\mathcal{D}$ 
represents its corresponding real-valued depth map. 

For learning $F$ we consider a deep model made of two main building blocks (Fig.~\ref{framework}). The first component is a CNN architecture with 
a set of intermediate side outputs $\mathcal{S}=\{ \vect{s}_{l} \}_{l=1}^{L}$, $\vect{s}_{l} \in R^N$, produced from $L$ different layers 
with a mapping function $f_s(\vect{r}; \mathbf{\Theta}, \boldsymbol{\theta}_{l}) \rightarrow \vect{s}_{l}$. 
For simplicity, we denote with $\mathbf{\Theta}$ the set of all network layer parameters and with $\boldsymbol{\theta}_{l}$ the parameters of 
the network branch producing the side output associated to the {$l$-th} layer (see Section \ref{setup} for details of our implementation). 
In the following we denote this network as the front-end CNN.

The second component of our model is a fusion block. As shown in previous works \cite{long2015fully,bertasius2015deepedge,xie2015holistically}, 
features generated from different 
CNN layers capture complementary information. The main idea behind the proposed fusion block
is to use CRFs to effectively integrate the side output maps of our front-end CNN 
for robust depth prediction. Our approach develops from the intuition that these representations 
can be combined within a sequential framework, 
\ie performing depth estimation at a certain scale
and then refining the obtained estimates in the subsequent level.
Specifically, we introduce and compare two different multi-scale models, both based on CRFs, and corresponding to two different version of the
fusion block. 
The first model is based on a \textbf{single multi-scale CRFs}, which integrates information
available from different scales and simultaneously enforces smoothness constraints between the estimated depth values of
neighboring pixels and neighboring scales. The second model implements a \textbf{cascade of scale-specific CRFs}: at each scale $l$
a CRF is employed to recover the depth information from side output maps $\mathbf{s}_l$ and the 
outputs of each CRF model are
used as additional observations for the subsequent model.
In Section \ref{sec:HCRF} we describe the two models in details, while in Section \ref{mean-field}
we show how they can be implemented as sequential deep networks by 
stacking several elementary blocks. We call these blocks C-MF blocks, as they implement Mean Field updates for Continuous CRFs (Fig.~\ref{framework}).

\subsection{Fusing side outputs with continuous CRFs}
We now describe the proposed CRF-based models for fusing multi-scale representations. \vspace{-3mm}
\paragraph{Multi-scale CRFs.} 
\label{sec:HCRF}
Given an $LN$-dimensional vector $\hat{\mathbf{s}}$ obtained by concatenating the side output 
{score maps} $\{\vect{s}_{1},\dots,\vect{s}_{L}\}$ and 
an $LN$-dimensional vector $\vect{d}$ of real-valued output variables, 
we define a CRF modeling the conditional distribution:
\begin{equation}
\setlength{\abovedisplayskip}{2pt}
\setlength{\belowdisplayskip}{2pt}
  P(\vect{d}|\hat{\mathbf{s}}) = \frac{1}{Z(\hat{\mathbf{s}})} \exp \{-E(\vect{d},\hat{\mathbf{s}})\}
\end{equation}
where $Z(\hat{\mathbf{s}})=\int_{\vect{d}} \exp {-E(\vect{d},\hat{\mathbf{s}})} {d\vect{d}}$ is the partition function.
The energy function is defined as:
\begin{equation}
\setlength{\abovedisplayskip}{2pt}
\setlength{\belowdisplayskip}{2pt}
E(\vect{d}, \hat{\mathbf{s}})  =  \sum_{i=1}^{N} \sum_{l=1}^{L} \phi(d_i^l, \hat{\mathbf{s}}) + \sum_{i, j} \sum_{l, k} 
 \psi(d_i^l, d_j^k)  \label{energy}
\end{equation}
and $d_i^l$ indicates the hidden variable associated to scale $l$ and pixel $i$.
The first term is the sum of quadratic unary terms defined as:
\begin{equation}
\setlength{\abovedisplayskip}{2pt}
\setlength{\belowdisplayskip}{2pt}
\phi(d_i^l, \hat{\mathbf{s}}) = \big (d_i^l - s_i^l \big )^2
\end{equation}
where $s^l_i$ is the regressed depth value at pixel $i$ and scale $l$ obtained with $f_s(\vect{r}; \mathbf{\Theta}, \boldsymbol{\theta}_{l})$.
The second term is the sum of pairwise potentials describing the relationship 
between pairs of hidden variables $d_i^l$ and $d_j^k$ and is defined as follows:
\begin{equation}
\setlength{\abovedisplayskip}{2pt}
\setlength{\belowdisplayskip}{2pt}
\psi(d_i^l, d_i^k) =  \sum_{m=1}^{M} \beta_m w_m(i,j,l,k,\vect{r})(d_i^l-d_j^k)^2
\end{equation}
where $w_m(i,j,l,k,\vect{r})$ is a weight which specifies the relationship between the estimated depth
of the pixels $i$ and $j$ at scale $l$ and $k$, respectively.

To perform inference we rely on mean-field approximation, \ie 
$Q(\vect{d}|\hat{\mathbf{s}})=\prod_{i=1}^N\prod_{l=1}^L Q_{i,l}(d_i^l|\hat{\mathbf{s}})$. 
Following \cite{ristovski2013continuous}, by considering
$J_{i,l}=\log Q_{i,l}(d_i^l|\hat{\mathbf{s}})$ and rearranging its expression into an exponential form, the following
mean-field updates can be derived:
\begin{equation}
\setlength{\abovedisplayskip}{2pt}
\setlength{\belowdisplayskip}{2pt}
\gamma_{i,l} = 2 \big(1+2\displaystyle\sum_{m=1}^{M}\beta_m \sum_{k} \sum_{j, i} w_m(i,j,l,k,\vect{r})\big)
\label{sigma}
\end{equation}
\begin{equation}
\setlength{\abovedisplayskip}{2pt}
\setlength{\belowdisplayskip}{2pt}
\mu_{i,l} = \frac{2}{\gamma_{i,l}} \big(s_i^l + 2\sum_{m=1}^{M}\beta_m \sum_{k} \sum_{j, i} w_m(i,j,l,k,\vect{r}) \mu_{j,k} \big)
\label{mu}
\end{equation}
To define the weights $w_m(i,j,l,k,\vect{r})$ we introduce the following assumptions. First, we assume that the estimated depth at scale $l$ only
depends on the depth estimated at previous scale. 
Second, for relating pixels at the same and at previous scale, we set
weights depending on $m$ Gaussian kernels $K_m^{ij}=\exp\big(-\frac{\|\mathbf{h}_i^m-\mathbf{h}_j^m\|^2_2}{2\theta_m^2}\big)$. Here,
$\vect{h}_i^m$ and $\vect{h}_j^m$ indicate some features 
derived from the input image $\vect{r}$ for pixels $i$ and $j$. $\theta_m$ are user defined parameters. 
Following previous works \cite{koltun2011efficient}, we use pixel position 
and color values as features, leading to two Gaussian kernels (\ie~an appearance and a smoothness kernel)
for modeling dependencies of pixels at scale $l$ and other two for relating pixels at neighboring scales. 
Under these assumptions, the mean-field updates (\ref{sigma}) and (\ref{mu}) can be rewritten as:
\begin{equation}
\setlength{\abovedisplayskip}{2pt}
\setlength{\belowdisplayskip}{2pt}
\gamma_{i,l} = 2 \big(1+2 \sum_{m=1}^{2} \beta_m \sum_{j\neq i}  K_m^{ij} + 2\sum_{m=3}^{4} \beta_m 
\sum_{j, i} K_m^{ij}  \big)
\label{gamma2}
\end{equation}
\setlength{\abovedisplayskip}{2pt}
\setlength{\belowdisplayskip}{2pt}
\begin{equation}
\begin{aligned}
&\mu_{i,l} =& \frac{2}{\gamma_{i,l}} \big(s_i^l + 2\sum_{m=1}^{2}\beta_m \sum_{j \neq i} K_m^{ij} \mu_{j,l}, \\
&&+ 2\sum_{m=3}^{4} \beta_m \sum_{j, i} K_m^{ij} \mu_{j,l-1} \big) 
\end{aligned}
\label{mu2}
\end{equation}
Given a new test image, the optimal $\tilde{\vect{d}}$ can be computed
maximizing the log conditional probability \cite{ristovski2013continuous}, \ie 
$\tilde{\vect{d}} = \mr{arg}\max_{\vect{d}} \log (Q(\vect{d}|\vect{S}))$,
where $\tilde{\vect{d}} = [\mu_{1,1},...,\mu_{N,L}]$ is a vector of the $LN$ mean values associated to $Q(\vect{d}|\hat{\mathbf{s}})$.
We take the estimated variables at the finer scale $L$ (\ie $\mu_{N,1},...,\mu_{N,L}$) as our predicted depth map $\mathbf{d}^\star$. 
\vspace{-3mm}

\paragraph{Cascade CRFs.}
The cascade model is based on a set of $L$ CRF models, each one associated to a specific scale $l$, which are
progressively stacked such that the estimated depth at previous scale can be used to define the features of the CRF model
in the following level.
Each CRF is used to compute the output vector $\vect{d}^l$ and it is 
constructed considering the side output representations $\vect{s}^l$ and the estimated depth at the previous
step $\tilde{\vect{d}}^{l-1}$ as observed variables, \ie $\vect{o}^l=[\vect{s}^l, \tilde{\vect{d}}^{l-1}]$.
The associated energy function is defined as:
\begin{equation}
\setlength{\abovedisplayskip}{2pt}
\setlength{\belowdisplayskip}{2pt}
E(\vect{d}^l, \vect{o}^l) = \sum_{i=1}^{N} \phi(d_i^l, \vect{o}^l) + \sum_{i \neq j} \psi(d_i^l, d_j^l).
\end{equation}
The unary and pairwise terms can be defined analogously to the unified model.
In particular the unary term, reflecting the similarity between the observation $o^i_l$ and the hidden depth value $d_i^l$, is:
\begin{equation}
\setlength{\abovedisplayskip}{2pt}
\setlength{\belowdisplayskip}{2pt}
\phi(y_i^l, \vect{o}^l) = \big (d_i^l - o_i^l \big )^2
\end{equation}
where $o_i^l$ is obtained combining the regressed depth from side outputs $\vect{s}^l$ and 
the map $\vect{d}^{l-1}$ estimated by CRF at previous scale. In our implementation we simply consider $o_i^l=s_i^l+\tilde{d}_i^{l-1}$, but other
strategies can be also considered.
The pairwise potentials, used to force neighboring pixels with similar appearance to have 
close depth values, are:
\begin{equation}
\setlength{\abovedisplayskip}{2pt}
\setlength{\belowdisplayskip}{2pt}
\psi(d_i^l, d_j^l) = \sum_{m=1}^{M}\beta_m K_m^{ij}(d_i^l-d_j^l)^2
\end{equation}
where we consider $M=2$ Gaussian kernels, one for appearance features, the other accounting for pixel positions.
Similarly to the multi-scale model, under mean-field approximation, the following updates can be derived:
\begin{equation}
\setlength{\abovedisplayskip}{2pt}
\setlength{\belowdisplayskip}{2pt}
\gamma_{i,l} = 2 \big(1+2 \sum_{m=1}^{M} \beta_m \sum_{j\neq i}  K_m^{ij}\big)
\label{gamma1}
\end{equation}
\begin{equation}
\begin{aligned}
\setlength{\abovedisplayskip}{2pt}
\setlength{\belowdisplayskip}{2pt}
&\mu_{i,l} =& \frac{2}{\gamma_{i,l}} \big(o_i^l + 2\sum_{m=1}^{M}\beta_m \sum_{j \neq i} K_m^{ij} \mu_{j,l}  \big)
\label{mu1}
\end{aligned}
\end{equation}
At test time, we use the estimated variables corresponding to the CRF model
of the finer scale $L$ as our predicted depth map $\mathbf{d}^\star$.

\subsection{Multi-scale models as sequential deep networks} \label{mean-field}
In this section, we describe how the two proposed CRFs-based models can be implemented
as sequential deep networks, enabling end-to-end training of our whole network model (front-end CNN and fusion module).
We first show how the mean-field iterations derived for the multi-scale and the cascade
models can be implemented by defining a common structure, the C-MF block, {consisting into a stack of CNN layers}. Then, we present the resulting sequential network structures and detail the training phase.
\vspace{-0.3cm}
\paragraph{C-MF: a common {CNN} implementation for two models.} 
By analyzing the two proposed CRF models, we can observe that 
the mean-field updates derived for the cascade and for the multi-scale models
share common terms. As stated above, the main difference between the two is the way 
the estimated depth at previous scale is handled at the current scale. In the multi-scale CRFs, 
the relationship among neighboring scales is modeled in the hidden variable space, while in the cascade CRFs the depth estimated at previous scale acts as an observed variable.



\begin{figure}[!t]
\centering
\includegraphics[width=3.5in]{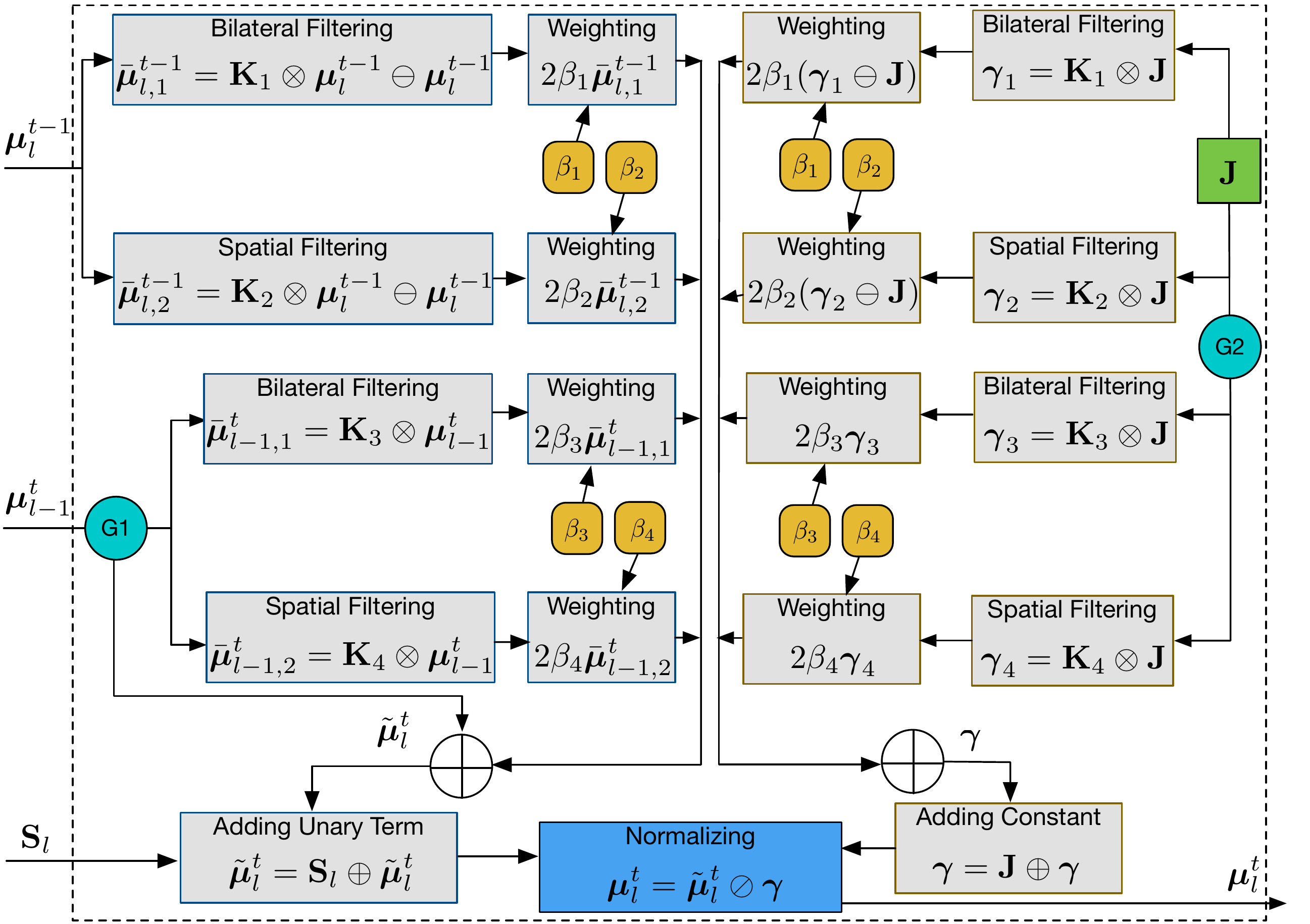} 
\caption{The proposed C-MF block. $\mathbf{J}$ represents a $W\times H$ matrix with all elements equal to one. 
The symbols $\oplus$, $\ominus$, $\oslash$ and $\otimes$ indicate element-wise addition, subtraction, division and Gaussian convolution, respectively.}
\label{fig:HCRF-MF}
\vspace{-0.3cm}
\end{figure}

Starting from this observation, in this section we show how the computation of Eq.~(\ref{mu2}) and (\ref{mu1}) can be implemented 
with a common structure. 
Figure~\ref{fig:HCRF-MF} describes in details these computations. In the following, for the sake of
clarity, we introduce matrices. Let ${\mathbf{S}}_l \in \mathbb{R}^{W\times H}$ be the matrix obtained rearranging 
the $N=WH$ pixels corresponding to the side outputs vector $\mathbf{s}_l$ and $\boldsymbol{\mu}_l^t \in \mathbb{R}^{W\times H}$ 
the matrix of the estimated output variables associated to scale $l$ and mean field iteration $t$.
To implement the multi-scale model at each iteration $t$, $\boldsymbol{\mu}_l^{t-1}$ and $\boldsymbol{\mu}_{l-1}^t$ are convolved 
by two Gaussian kernels. Following \cite{koltun2011efficient}, we use a spatial and a bilateral kernel. As Gaussian convolutions 
represent the computational bottleneck in the mean-field iterations, 
we adopt the permutohedral lattice implementation~\cite{adams2010fast} to approximate
filter response calculation reducing computational cost from quadratic to linear~\cite{ristovski2013continuous}.
The weighing of the parameters $\beta_m$ is performed as a convolution with a $1 \times 1$ filter.
Then, the outputs are combined and are added to the side output maps $\mathbf{S}_l$. 
Finally, a normalization step follows, corresponding to the calculation of (\ref{gamma2}). The normalization matrix 
$\boldsymbol{\gamma} \in \mathbb{R}^{W \times H}$ is also computed by considering Gaussian kernels convolutions and 
weighting with parameters $\beta_m$. It is worth noting that the normalization step in our mean-field updates for 
continuous CRFs is substantially different from that of discrete CRFs~\cite{zheng2015conditional} based on a softmax function. 

In the cascade CRF model, differently from the multi-scale CRF, $\boldsymbol{\mu}_{l-1}^t$ acts as an observed variable. To design a common C-MF block among the two models, we introduce two gate functions G1 and G2 (Fig.~\ref{fig:HCRF-MF}) controlling the computing flow and allowing to easily switch between the two approaches. Both gate functions accept a user-defined boolean parameter (here 1 corresponds to the multi-scale CRF and 0 to the cascade model). Specifically, if G1 is equal to 1, the gate function 
G1 passes $\boldsymbol{\mu}_{l-1}^{t}$ to the Gaussian filtering block, otherwise to the addition block with unary term. 
Similarly, G2 controls the computation of the normalization terms and switches between the computation of (\ref{gamma2}) 
and (\ref{gamma1}). Importantly, for each step in the C-MF block we implement the calculation of error differentials for 
the back-propogation as in~\cite{zheng2015conditional}. For optimizing the CRF parameters, similar to~\cite{zheng2015conditional}, the 
bandwidth values $\theta_m$ are fixed and we implement the differential computation for the weights of Gaussian kernels $\beta_m$.
In this way $\beta_{m}$ are learned automatically with back-propagation.
\vspace{-0.5cm}

\begin{figure*}[!t]
\centering
\includegraphics[width=3.5in]{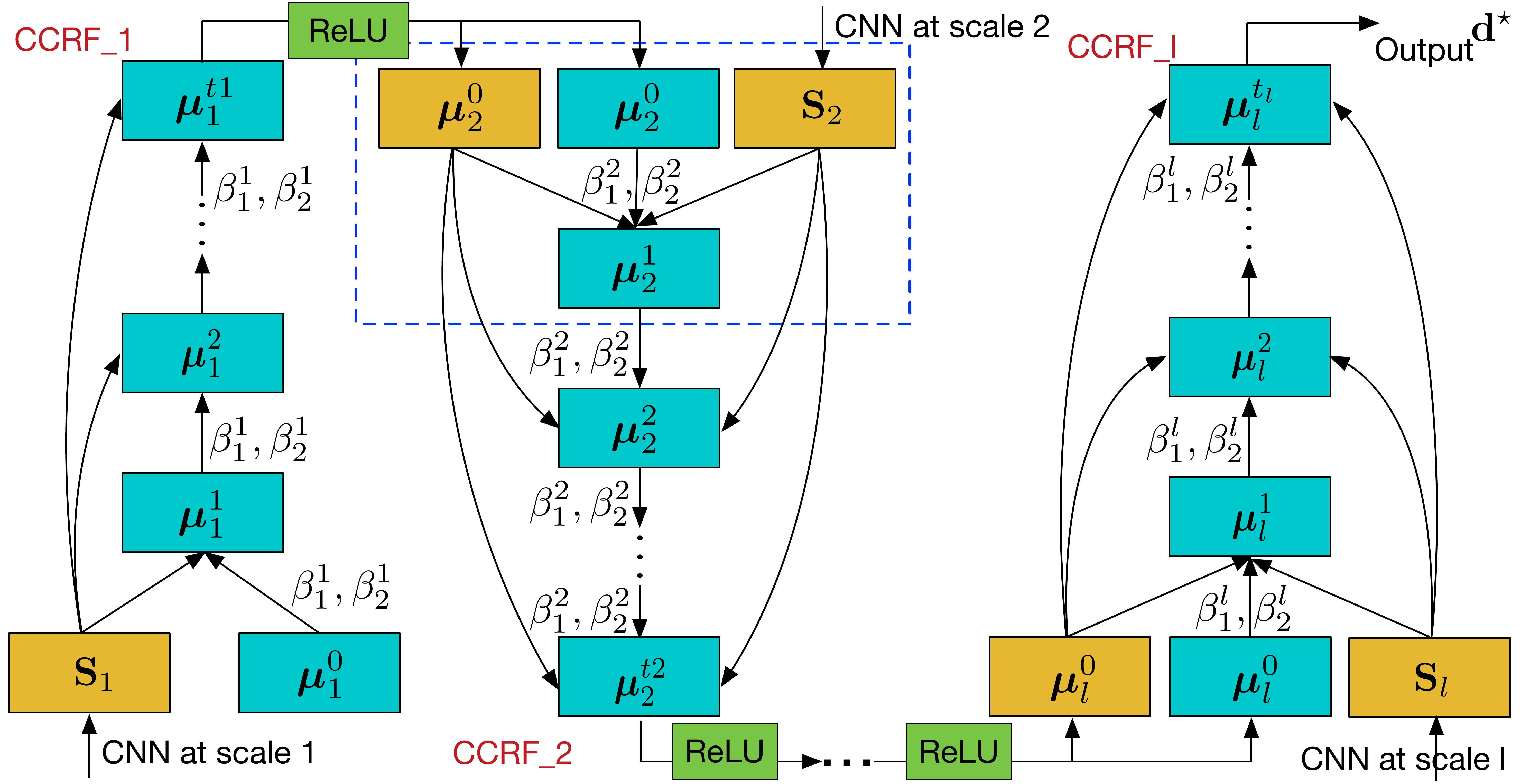} \ \ \ \ \ \ \ \ \ \ 
\includegraphics[width=2.75in]{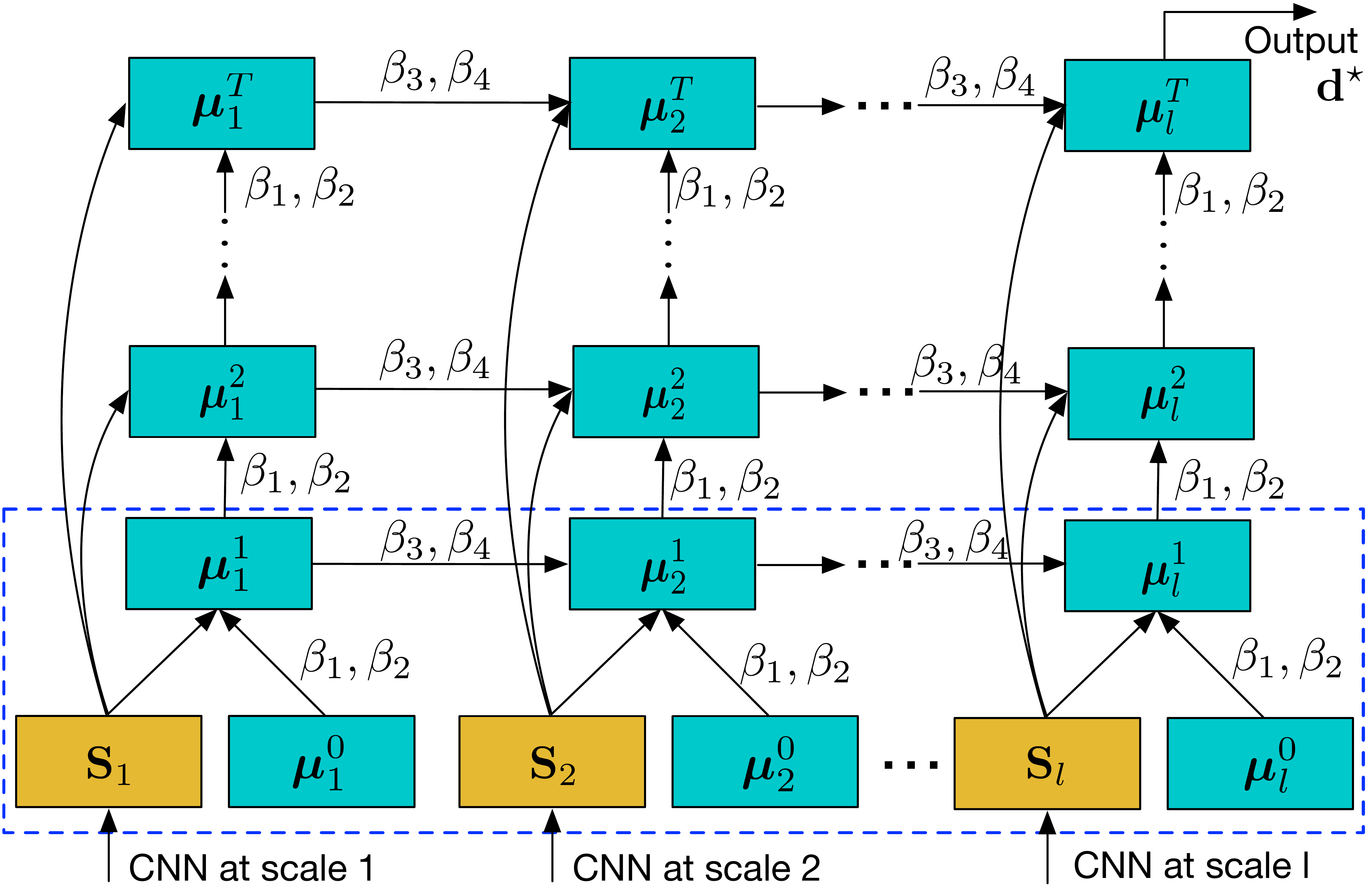} 
\caption{The proposed cascade (left) and multi-scale (right) models as a sequential deep networks. The blue and yellow boxes indicate the
estimated variables and observations, respectively. The parameters ${\beta}_m$ are used for mean-field updates. 
As in the cascade model parameters are not shared among different CRFs, we use the notation 
${\beta}_1^l, {\beta}_2^l$ to denote parameters associated to the $l$-th scale.
}
\label{sccnf}
\vspace{-0.5cm}
\end{figure*}
\paragraph{From mean-field updates to sequential deep networks.}
Figure~\ref{sccnf} illustrates the implementation of the proposed two CRF-based models using the C-MF block described above. 
In the figure, each blue box is associated to a mean-field iteration. 
The cascade model (Fig.~\ref{sccnf}-left) consists of $L$ single-scale CRFs. At the $l$-th scale, 
$t_l$ mean-field iterations are performed and then the estimated outputs are passed to the CRF model of the 
subsequent scale after a Rectified Linear Unit (ReLU) operation. To implement a single CRF, we stack $t_l$ C-MF blocks and 
make them share the parameters, while we learn different parameters for different CRFs. For the multi-scale model, one full 
mean-field updates involves $L$ scales simultaneously, obtained by combining $L$ C-MF blocks. We further stack $T$ 
iterations for learning and inference. The parameters corresponding to different scales and different 
mean-field iterations are shared. In this way, by using the common C-MF layer, we implement the two proposed CRFs models as 
deep sequential networks enabling end-to-end training with the front-end network.
\vspace{-0.45cm}
\paragraph{Training the whole network.}\label{learning}
We train the network using a two phase scheme. In the first phase (pretraining), the 
parameters $\mathbf{\Theta}$ and $\{\boldsymbol{\theta}_l\}_{l=1}^L$ 
of the front-end network are learned
by minimizing the sum of $L$ distinct side losses as in \cite{xie2015holistically}, corresponding to $L$ side 
outputs. We use a square loss over $Q$ training samples: $\mathcal{L}_P=\sum_{l=1}^L  \sum_{i=1}^{Q}\|\vect{s}_{l,i} - \bar{\vect{d}}_i\|_2^2$. 
In the second phase (fine tuning), we initialize the front-end network with the learned parameters in the first phase, and 
jointly fine-tune with the proposed multi-scale CRF models to compute the optimal value of the 
parameters $\mathbf{\Theta}, \{\boldsymbol{\theta}_l\}_{l=1}^L$ and $\boldsymbol{\beta}$, with 
$\boldsymbol{\beta}=\{\beta_m\}_{m=1}^M$. The entire network is learned with Stochastic Gradient Descent (SGD) 
by minimizing a square loss $\mathcal{L}_F=\sum_{i=1}^{Q}\|F(\vect{r}_i; \mathbf{\Theta},\boldsymbol{\theta}_l,\boldsymbol{\beta}) - \vect{d}_i^l\|_2^2$.  

%% file: experiments_e.tex
To demonstrate the effectiveness of the proposed multi-scale CRF models for monocular depth prediction,
we performed experiments on two publicly available datasets: 
the NYU Depth V2 ~\cite{silberman2012indoor} and the Make3D~\cite{saxena2005learning} datasets. In the
following we describe the details of our evaluation.
\subsection{Experimental Setup}
\label{setup}
\textbf{Datasets.} The \textbf{NYU Depth V2} dataset~\cite{silberman2012indoor} contains 120K unique pairs of RGB and depth images captured with a 
Microsoft Kinect. The datasets consists of 249 scenes for training and 215 scenes for testing. The images
have a resolution of $640 \times 480$. To speed up the training phase, following previous works~\cite{liu2015deep,zhuo2015indoor}
we consider only a small subset of images. This subset has 1449 aligned RGB-depth pairs:
795 pairs are used for training, 654 for testing. Following~\cite{eigen2014depth}, we perform data augmentation for 
the training samples. The RGB and depth images are scaled with a ratio $\rho \in \{1, 1.2, 1.5\}$ and the depths are divided by $\rho$. Additionally,
we horizontally flip all the samples and bilinearly down-sample them to $320 \times 240$ pixels. The data augmentation phase produces 
4770 training pairs in total. 

The \textbf{Make3D} dataset~\cite{saxena2005learning} contains 534 RGB-depth pairs, split into 400 pairs for training and 134 for testing. 
We resize all the images to a resolution of $460 \times 345$ as done in~\cite{liu2014discrete} to preserve the aspect ratio 
of the original images. We adopted the same data augmentation scheme used for NYU Depth V2 dataset but,
for $\rho=\{1.2,1.5\}$ we generate two samples each, obtaining 4K training samples. 

\textbf{Front-end CNN Architectures.} 
To study the influence of the frond-end CNN, we consider several network architectures 
including: (i) AlexNet~\cite{krizhevsky2012imagenet}, (ii) VGG16~\cite{simonyan2014very}, 
(iii) an encoder-decoder network derived from VGG (VGG-ED)~\cite{badrinarayanan2015segnet}, 
(iv) VGG Convolution-Deconvolution (VGG-CD)~\cite{noh2015learning}, and (v) ResNet50~\cite{he2015deep}. For AlexNet, VGG16 and ResNet50, 
we obtain the side outputs from different convolutional blocks in which each convolutional layer outputs feature maps with the same size 
using a similar scheme as in~\cite{xie2015holistically}.
The number of side-outputs is 5, 5 and 4 for AlexNet, VGG16 and ResNet50, respectively. As VGG-ED and VGG-CD have been widely used for pixel-level prediction tasks, we also consider them in our analysis. Both VGG-ED and VGG-CD have a symmetric structure, and we use the corresponding part of VGG16 for their encoder/convolutional block. Five side-outputs are then extracted from the convolutional blocks of the decoder/deconvolutional part.

\textbf{Evaluation Metrics.}
Following previous works~\cite{eigen2015predicting,eigen2014depth,wang2015towards}, we adopt the following evaluation metrics 
to quantitatively assess the performance of our depth prediction model. Specifically, we consider: (i) mean relative error (rel): 
\( \frac{1}{N}\sum_i\frac{|\bar{d}_i - d_i^\star|}{d_i^\star} \); (ii) root mean squared error (rms): 
\( \sqrt{\frac{1}{N}\sum_{i}(\bar{d}_i - d_i^\star)^2} \); 
(iii) mean log10 error (log10): \( \frac{1}{N}\sum_i \Vert \log_{10}(\bar{d}_i) - \log_{10}(d_i^\star) \Vert \) and
(iv) accuracy with threshold $t$: percentage (\%) of $d_i^\star$ subject to $\max (\frac{d_i^\star}{\bar{d}_i}, \frac{\bar{d}_i}{d_i^\star}) = 
\delta < t~(t \in [1.25, 1.25^2, 1.25^3])$. 

\textbf{Implementation Details.}
We implement the proposed deep model using the popular Caffe framework~\cite{jia2014caffe} on a single Nvidia Tesla K80 GPU with 12 GB memory. 
As described in Section~\ref{learning}, training consists of a pretraining and a fine tuning phase. In the first phase, we train the front-end CNN with parameters initialized with the corresponding ImageNet pretrained models. For AlexNet, VGG16, VGG-ED and VGG-CD, the batch size is set to 12 and for ResNet50 to 8. 
The learning rate is initialized at $10^{-11}$ and decreases by 10 times around every 50 epochs. 80 epochs are performed for pretraining in total. 
The momentum and the weight decay are set to 0.9 and 0.0005, respectively. 
{When the pretraining is finished, we connect all the side outputs of the front-end CNN to our CRFs-based multi-scale deep models 
for end-to-end training of the whole network.} In this phase, the batch size is reduced to 6 and a fixed 
learning rate of $10^{-12}$ is used. The same parameters of the pre-training phase are 
used for momentum and weight decay. The bandwidth weights for the Gaussian kernels are obtained 
through cross validation. {The number of mean-field iterations is set to 5 for efficient training for both 
the cascade CRFs and multi-scale CRFs. We do not observe significant improvement using more than 5 iterations.} 
Training the whole network takes around 25 hours on the Make3D dataset and $\sim 31$ hours on the NYU v2 dataset. 

\subsection{Experimental Results}
\begin{table}
\centering
\huge
\resizebox{0.49\textwidth}{!} {
\begin{tabular}{l|ccc|ccc}
\toprule
\multirow{2}{*}{Method} & \multicolumn{3}{c|}{\tabincell{c}{Error (lower is better)}} & \multicolumn{3}{c}{\tabincell{c}{Accuracy (higher is better)}} \\\cline{2-7}
                                      & rel & log10 & rms & $\delta < 1.25$ & $\delta < 1.25^2$ & $\delta < 1.25^3$ \\\hline
HED~\cite{xie2015holistically}      & 0.185 & 0.077    &    0.723   & 0.678& 0.918&  0.980\\
Hypercolumn~\cite{hariharan2015hypercolumns}      &0.189 & 0.080 & 0.730 & 0.667&0.911&  0.978\\
CRF             & 0.193& 0.082& 0.742& 0.662 & 0.909 & 0.976\\ \hline
Ours (single-scale)                & 0.187     &    0.079   &    0.727  & 0.674& 0.916&  0.980 \\
Ours - Cascade (3-s)              &0.176   &    0.074    &    0.695  & 0.689 & 0.920&  0.980     \\
Ours - Cascade (5-s)            &0.169  & 0.071& 0.673  & 0.698 & 0.923 &  \textbf{0.981}     \\
Ours - Multi-scale (3-s)           &  0.172 & 0.072   &  0.683   &  0.691   &  0.922   &  \textbf{0.981} \\ 
Ours - Multi-scale (5-s)           &  \textbf{0.163} &  \textbf{0.069} &   \textbf{0.655}  & \textbf{ 0.706 }  &\textbf{ 0.925}    &   \textbf{0.981}\\ 
\bottomrule                            
\end{tabular}
}
\caption{NYU Depth V2 dataset. Comparison of different multi-scale fusion schemes. 3-s, 5-s denote 3 and 5 scales respectively.}
\label{comparison}
\vspace{-0.3cm}
\end{table}

\begin{table}
\centering
\Large
\resizebox{0.49\textwidth}{!} {
\begin{tabular}{l|ccc|ccc}
\toprule
\multirow{2}{*}{Method} & \multicolumn{3}{c|}{\tabincell{c}{Error (lower is better)}} & \multicolumn{3}{c}{\tabincell{c}{Accuracy (higher is better)}} \\\cline{2-7}
                                      & rel & log10 & rms & $\delta < 1.25$ & $\delta < 1.25^2$ & $\delta < 1.25^3$ \\\hline
Outer $\rightarrow$ Inner        & 0.175  &  0.072 & 0.688 & 0.689 & 0.919 & 0.979 \\
Inner $\rightarrow$ Outer        & \textbf{0.169}  & \textbf{0.071}& \textbf{0.673} & \textbf{0.698} & \textbf{0.923} & \textbf{0.981} \\
\bottomrule                            
\end{tabular}
}
\caption{NYU Depth V2 dataset. Comparison between the proposed model and the associated pretrained network architectures.}
\label{cnn_order}
\vspace{-0.3cm}
\end{table}

\textbf{Analysis of different multi-scale fusion methods.} In the first series of experiments we consider
the NYU Depth V2 dataset. We evaluate the proposed CRF-based models 
and compare them with other methods for fusing multi-scale CNN representations. 
Specifically, we consider: (i) the HED method in~\cite{xie2015holistically}, where the sum of multiple side output losses
is jointly minimized with a fusion loss (we use the square loss, rather than the cross-entropy, as our problem involves
continuous variables), (ii) Hypercolumn~\cite{hariharan2015hypercolumns}, where 
multiple score maps are concatenated and (iii) a CRF applied on the prediction of the front-end network (last layer)
\textit{a posteriori} (no end-to-end training).
In these experiments we consider VGG-CD as front-end CNN.

The results of our comparison are shown in Table~\ref{comparison}. It is evident that 
with our CRFs-based models more accurate depth maps can be obtained, confirming our idea
that integrating complementary information derived from CNN side output maps within a graphical model framework
is more effective than traditional fusion schemes. 
{The table also compares the proposed cascade and multi-scale models. As expected, the multi-scale model produces more accurate depth maps, 
at the price of an increased computational cost.}
Finally, we analyze the impact of adopting multiple scales and compare our complete models (5 scales)
with their version when only a single and three side output layers are used. It is evident that the performance improves by increasing the
number of scales. 

As the proposed models are based on the idea of progressively refining the obtained prediction results from previous layers, we 
also analyze the influence of the stacking order on the performance of the cascade model (Table~\ref{cnn_order}). We compare two different schemes: 
the first indicating that the cascade model operates from the inner to the outer layers and the other representing the reverse order.
Our results confirm the validity of our original assumption: a coarse to fine approach leads to 
more accurate depth maps.

\textbf{Evaluation of different front-end deep architectures.} As discussed above, the proposed multi-scale fusion models are
general and different deep neural architectures can be employed in the front end network. In this section,
we evaluate the impact of this choice on the depth estimation performance. The results of our analysis are shown in
Table~\ref{cnn_arch}, where we consider both the case of pretrained model (\ie only side losses are employed but not CRF models), indicated with P,
and the fine-tuned model with the cascade CRFs (CRF). Similar results
are obtained in the case of the multi-scale CRF. As expected, in both cases deeper models produced more accurate predictions and ResNet50 outperforms other models. Moreover, VGG-CD is slightly better than VGG-ED, and both these models outperforms VGG16.
Importantly, for all considered networks there is a significant increase in performance when applying the proposed CRF-based models.

\begin{figure*}[!t]
\centering
\includegraphics[width=6.42in]{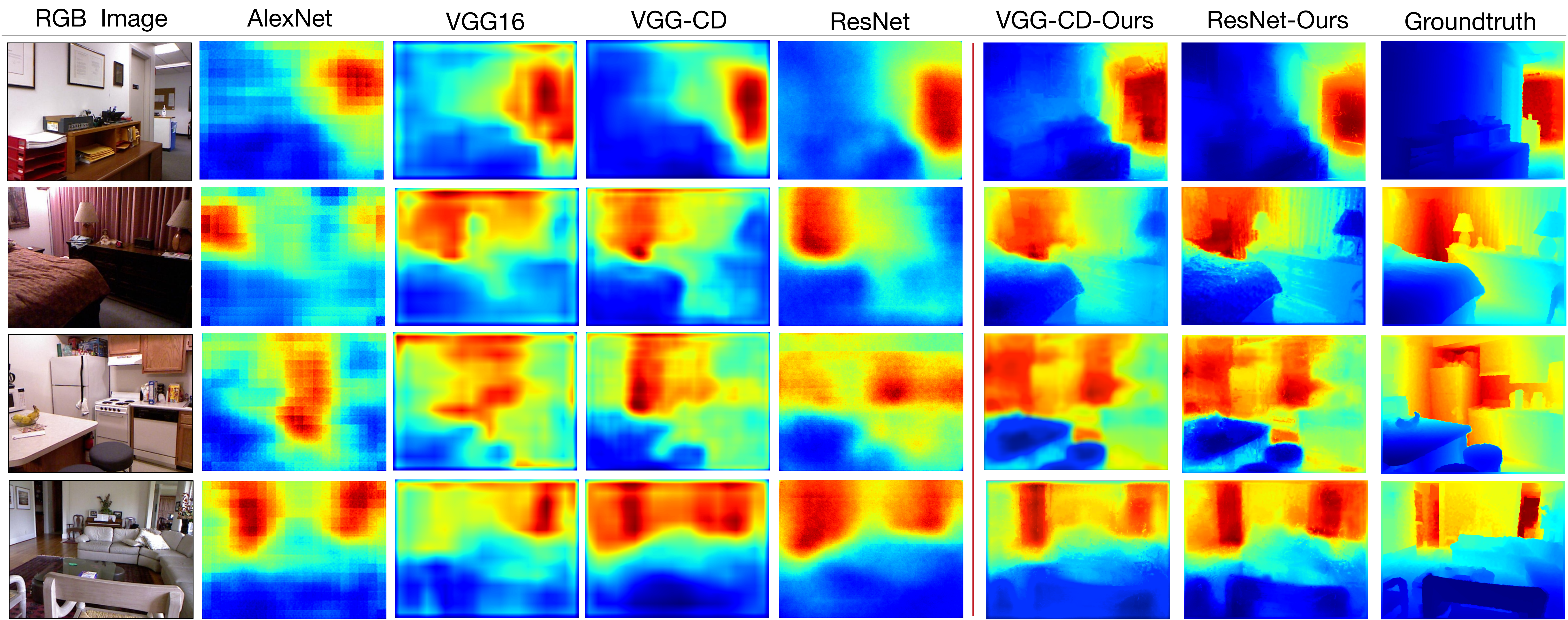} 
\caption{Examples of depth prediction results on the NYU v2 dataset. Different network architectures are compared.}
\label{nyu2examples}
\vspace{-0.3cm}
\end{figure*}

Figure~\ref{nyu2examples} depicts some examples of predicted depth maps on the NYU Depth V2 dataset. As shown in the figure, the proposed approach 
is able to generate robust depth predictions. By comparing the reconstructed depth images
obtained with pretrained models (\eg using VGG-CD and ResNet50 as front-end networks) with those computed with our 
models, it is clear that our multi-scale approach significantly improves prediction accuracy.

\begin{table}
\centering
\Large
\resizebox{0.49\textwidth}{!} {
\begin{tabular}{l|ccc|ccc}
\toprule
\multirow{2}{*}{\tabincell{c}{Network \\Architecture}} & \multicolumn{3}{c|}{\tabincell{c}{Error (lower is better)}} & \multicolumn{3}{c}{\tabincell{c}{Accuracy (higher is better)}} \\\cline{2-7}
                                      & rel & log10 & rms & $\delta < 1.25$ & $\delta < 1.25^2$ & $\delta < 1.25^3$ \\\hline
AlexNet (P)       & 0.265 &0.120&0.945 & 0.544 & 0.835 & 0.948 \\
VGG16 (P)                  &0.228  &0.104&0.836& 0.596 & 0.863 & 0.954 \\
VGG-ED (P)               & 0.208 & 0.089 &0.788& 0.645 & 0.906 & 0.978\\
VGG-CD (P)               & 0.203 & 0.087&0.774&  0.652 & 0.909& 0.979 \\
ResNet50 (P)             & 0.168& 0.072& 0.701& 0.741 & 0.932 & 0.981\\ \hline
AlexNet (CRF)               & 0.231 &  0.105     & 0.868 &0.591 & 0.859 & 0.952 \\
VGG16 (CRF)               &0.193    &0.092& 0.792    &   0.636     &     0.896    &    0.972     \\
VGG-ED (CRF)             &0.173    &0.073& 0.685    &   0.693     &     0.921    &    0.981     \\
VGG-CD (CRF)             & 0.169  & 0.071& 0.673    &   0.698     &     0.923    &  0.981 \\
ResNet50 (CRF)            & \textbf{0.143 } &\textbf{ 0.065}& \textbf{0.613}&\textbf{ 0.789} & \textbf{0.946} & \textbf{0.984} \\
\bottomrule                            
\end{tabular}
}
\caption{NYU Depth V2 dataset. Comparison between the proposed model and the associated pretrained network architectures.}
\label{cnn_arch}
\vspace{-0.3cm}
\end{table}

\begin{table}
\centering
\huge
\resizebox{1.05\linewidth}{!} {
\begin{tabular}{l|ccc|ccc}
\toprule
\multirow{2}{*}{Method} & \multicolumn{3}{c|}{\tabincell{c}{Error (lower is better)}} & \multicolumn{3}{c}{\tabincell{c}{Accuracy (higher is better)}} \\\cline{2-7}
                                      & rel & log10 & rms & $\delta < 1.25$ & $\delta < 1.25^2$ & $\delta < 1.25^3$ \\\hline
Karsch \etal~\cite{saxena2009make3d}                    & 0.349 &  -     & 1.214  &0.447 & 0.745 & 0.897 \\
Ladicky \etal~\cite{karsch2014depth}               &0.35    &0.131&1.20   &    -     &     -    &    -      \\
Liu \etal~\cite{liu2014discrete} & 0.335 & 0.127 & 1.06  &    -     &     -    &     -     \\
Ladicky \etal~\cite{ladicky2014pulling}               &   -       &    -    &    -     & 0.542& 0.829&  0.941 \\
Zhuo \etal~\cite{zhuo2015indoor}                       & 0.305 &0.122& 1.04  & 0.525& 0.838& 0.962 \\
Liu \etal~\cite{liu2015deep}                                 & 0.230 &0.095& 0.824& 0.614& 0.883&  0.975 \\
Wang \etal~\cite{wang2015towards}                   & 0.220 & 0.094&0.745& 0.605 & 0.890 & 0.970 \\
Eigen \etal~\cite{eigen2014depth}                      & 0.215  &  -      & 0.907& 0.611 &  0.887  &  0.971\\
Roi and Todorovic~\cite{roymonocular}                                   & 0.187  &  0.078& 0.744& - &  -  &  -\\
Eigen and Fergus~\cite{eigen2015predicting}               & 0.158  & -       & 0.641& 0.769 & 0.950 & 0.988 \\
Laina \etal~\cite{laina2016deeper}                     & 0.129  &0.056& \textbf{0.583}&0.801 & 0.950 & 0.986\\\hline 
Ours (ResNet50-4.7K)                                 & 0.143  & 0.065& 0.613   & 0.789 & 0.946 & 0.984 \\
Ours (ResNet50-95K)                                  & \textbf{0.121}  & \textbf{0.052}& 0.586 &\textbf{0.811}  & \textbf{0.954} &\textbf{0.987} \\
\bottomrule                           
\end{tabular}
}
\caption{NYU Depth V2 dataset: comparison with state of the art.}
\label{overall_nyu}
\vspace{-0.3cm}
\end{table}

\begin{table}[!t]
\LARGE
\centering
\resizebox{1\linewidth}{!} {
\begin{tabular}{l|ccc|ccc}
\toprule
\multirow{2}{*}{Method} & \multicolumn{3}{c|}{\tabincell{c}{C1 Error}} & \multicolumn{3}{c}{\tabincell{c}{C2 Error}} \\\cline{2-7}
                                      & rel & log10 & rms & rel & log10 & rms \\\hline
Karsch \etal~\cite{karsch2014depth}           &0.355  & 0.127 & 9.20 & 0.361   & 0.148 & 15.10 \\
Liu et al.~\cite{liu2014discrete} & 0.335 & 0.137 & 9.49 & 0.338 & 0.134 & 12.60 \\
Liu et al.~\cite{liu2015deep}      & 0.314    &0.119 & 8.60    & 0.307   &    0.125     &     12.89      \\
Li et al.~\cite{li2015depth}   &   0.278       &    0.092    &    7.19     & 0.279      & 0.102& 10.27 \\
Laina \etal~\cite{laina2016deeper} ($\ell_2$ loss)    &   0.223 & 0.089 &  4.89   & -       &    -     &  \\
Laina \etal~\cite{laina2016deeper} (Huber loss)    &   \textbf{0.176} & 0.072 &  4.46   & -       &    -     &  \\
\hline
Ours (ResNet50-Cascade)          &   0.213  & 0.082 &   4.67  & 0.221  & 4.79 & 8.81 \\
Ours (Resnet50-Multi-scale) &   0.206 & 0.076 &   4.51  & 0.212 & 4.71 & 8.73 \\
Ours (Resnet50-10K) &   0.184  &   \textbf{0.065}    &   \textbf{ 4.38}     & \textbf{0.198}  &   \textbf{4.53}  & \textbf{8.56} \\
\bottomrule                         
\end{tabular}
}
\caption{Make3D dataset: comparison with state of the art.}
\label{overall_make3d}
\vspace{-0.5cm}
\end{table}

\textbf{Comparison with state of the art.} We also compare our approach with state of the art methods on both datasets.
For previous works we directly report results taken from the original papers. Table~\ref{overall_nyu} shows 
the results of the comparison on the NYU Depth V2 dataset. 
For our approach we consider the cascade model and use two different training sets for pretraining: the small set of
4.7K pairs employed in all our experiments and a larger set of 95K images as in \cite{laina2016deeper}. Note that for fine tuning
we only use the small set. As shown in the table, our approach outperforms all baseline methods 
and it is the second best model when we use only 4.7K images. This is remarkable considering that,
for instance, in \cite{eigen2015predicting} 120K image pairs are used for training.

We also perform a comparison with state of the art on the Make3D dataset (Table~\ref{overall_make3d}).
Following~\cite{liu2014discrete}, the error metrics are computed in two different settings, \ie considering (C1) only 
the regions with ground-truth depth less than 70 and (C2) the entire image. 
It is clear that the proposed approach is significantly better than previous methods. 
In particular, comparing with Laina \etal~\cite{laina2016deeper}, the best performing method in the literature, 
it is evident that our approach, both in case of the cascade and the multi-scale models, outperforms \cite{laina2016deeper} by a 
significant margin when Laina \etal also adopt a square loss. It is worth noting that in \cite{laina2016deeper} a 
training set of 15K image pairs is considered, while we employ much less training samples. 
By increasing our training data (\ie $\sim 10$K in the pretraining phase), our multi-scale CRF model also outperforms \cite{laina2016deeper}
with Huber loss (log10 and rms metrics). Finally, it is very interesting to compare the proposed method with the 
approach in Liu \etal\cite{liu2015deep}, since in \cite{liu2015deep} a CRF model is also employed within a deep network trained end-to-end. 
Our method significantly outperforms \cite{liu2015deep} in terms of accuracy. Moreover, in \cite{liu2015deep} a time of 1.1sec is reported 
for performing inference on a test image but the time required by superpixels calculations is not taken into account. Oppositely, with 
our method computing the depth map for a single image takes about 1 sec in total.